\documentclass[10pt,twocolumn,a4paper]{article}

\usepackage[a4paper, total={180mm, 240mm}]{geometry}

\usepackage[utf8]{inputenc}

\usepackage{hyperref}

\usepackage{epsfig}
\usepackage{graphicx}
\usepackage{amsmath}
\usepackage{amssymb}
\usepackage{rotating}
\usepackage{lettrine}
\usepackage{flushend}
\usepackage{times}
\usepackage{natbib}

\def \ie {\emph{i.e.}}
\def \eg {\emph{e.g.}}
\def \etal {\emph{et al.}}
\newcommand{\myemph}[1]{\textbf{#1}}

\begin{document}

\title{An evaluation of large-scale methods \\ for image instance and class discovery}

\author{Matthijs Douze, Herv{\'e} J{\'e}gou, Jeff Johnson \\
Facebook AI Research \\
contact: matthijs@fb.com}

\date{}

\maketitle

\begin{abstract}

This paper aims at discovering meaningful subsets of related images from large image collections without annotations.
We search groups of images related at different levels of semantic, \ie, either instances or visual classes.
While k-means is usually considered as the gold standard for this task, we evaluate and show the interest of diffusion methods that have been neglected by the state of the art, such as the Markov Clustering algorithm.

We report results on the ImageNet and the Paris500k instance dataset, both enlarged with images from YFCC100M. We evaluate our methods with a labelling cost that reflects how much effort a human would require to correct the generated clusters.

Our analysis highlights several properties. First, when powered with an efficient GPU implementation, the cost of the discovery process is small compared to computing the image descriptors, even for collections as large as 100 million images.
Second, we show that descriptions selected for instance search improve the discovery of object classes. Third, the Markov Clustering technique consistently outperforms other methods; to our knowledge it has never been considered in this large scale scenario.

\end{abstract}

\section{Introduction}

\lettrine{L}{arge} collections of images are now prominent. The diversity of their visual content is high, and due to the ``long-tail'' issue well known by researchers working on text data, a few classes are very frequent, but the vast majority of the classes do not occur often. In the visual world we consider, it is hard to collect enough labelled data for most of the visual entities.
This is in contrast with the balanced and strongly supervised setting of ImageNet~\cite{DSLLF09}.

In our paper, we consider the problem of \myemph{visual discovery}. The task is to automatically suggest subsets of related images, without employing any label or tag. This differs from semi-supervised learning~\cite{FWT09}, where a fraction of the dataset is annotated beforehand with a pre-defined set of labels. It is also different from noisy supervision with unreliable hashtags, as in Joulin \etal~\cite{JMJV16}.
Most of the early work on discovery focused on instances~\cite{CM10a}, location recognition or city-level 3D reconstruction~\cite{ASSSS09,FGGJR10}, where the best methods are powered by spatial recognition, guaranteeing high matching performance by drastically reducing the rate of false positives. Such methods are not applicable to non-rigid instances or classes. Few studies have considered the problem of class discovery, which is harder to define from a user interest point of view, beyond classical clustering metrics like the square loss. %

We address a general discovery scenario, with an application in mind where we need to detect visually related images from a novel collection for the purpose of navigation, trend analysis or fast labelling. In this context, the user interest could be related to \myemph{categories} depicted in the collection but unseen at train time, or to \myemph{specific objects} such as paintings or locations.
For example, given a collection of landmark images, how can we determine that the user's interest is in distinguishing between Romanesque and Gothic architectures, or between the fa\c{c}ade of the Notre Dame cathedral and other buildings?
This problem is challenging because it addresses different levels of semantics, which are not necessarily well identified by a single kind of descriptor. For this purpose, we study recent candidate methods initially designed for instance recognition and image classification, namely R-MAC~\cite{TSJ16} and Resnet~\cite{kaiming16resnet}, and several discovery mechanisms based on kNN graphs and clustering. Our approach exploits dataset characteristics: if the dataset contains many Notre Dame images, then they will get a group of their own, otherwise they can be grouped with other Gothic cathedrals.

Our paper makes the following main contributions:
\begin{itemize}
\itemsep0em
\item We propose an \myemph{evaluation protocol} for the proposed discovery task, which accounts for different semantic levels and is extensible to arbitrarily large datasets using a distractor dataset. %
\item We evaluate the performance and scalability of \myemph{four clustering strategies}, namely k-means, agglomerative clustering, power-iterative clustering and an improved variant of the Markov Cluster Algorithm.
\item We show that when efficient CPU and GPU implementations of kNN search are used, diffusion methods can easily handle 10- to 100- million scale datasets, \ie, \myemph{one or two orders of magnitude larger} than the most accurate competing methods based on approximate k-means or diffusion, \eg, the works of Avrithis \etal~\cite{AKAE15} and Iscen \etal~\cite{iscen2017efficient,iscen2017fast}.
\item We apply Markov Clustering to this task, and show it significantly \myemph{outperforms k-means}, which is considered as a top-line in other approaches.
\end{itemize}

As a result of our study, we provide recommendations for the discovery task, and propose choices that will hopefully serve as baselines in future work on large-scale discovery.

The rest of this paper is organized as follows. After introducing related work in Section~\ref{sec:related}, we introduce the large-scale discovery strategy in Section~\ref{sec:method}. The experiments and evaluation are presented in Section~\ref{sec:experiments}. Section~\ref{sec:conclusion} concludes the paper.

\section{Related work}
\label{sec:related}

This section presents related work on visual discovery, associated with various problems like image description, classification and efficient clustering.
Note that typical descriptors employed for class and instance recognition are different. Even though these problems mainly differ by composition granularity, they are addressed by two distinct tasks and evaluation protocols in the literature, namely image classification and instance search/image search. We provide background references on these related tasks and cite relevant description schemes that we employ as input for our method. We also discuss prior art on discovery, including algorithms that aim to improve scalability.

\paragraph{Image descriptors for class and instance discovery.}

Traditionally, discovery~\cite{GWGL14,BSCL14,TSJ16} uses image description methods borrowed from image matching, in particular those based on keypoint indexing~\cite{L04,SZ03,PCISZ08,TAJ13}, with impressive results when fine-tuned for rigid objects, like buildings on the Oxford dataset~\cite{GARL16,RTC16}. For class discovery or semi-supervised labelling~\cite{FWT09}, semantic global descriptors like GIST~\cite{OT01} are preferred. Recently, classification performance has substantially improved with deep CNN architectures~\cite{SZ14,kaiming16resnet} which are therefore compelling choices for our purpose.

Weiler \& Fergus~\cite{ZF14} visualize the object classes corresponding to different activation levels of AlexNet and show that semantic levels correspond to layers. For networks trained on a dataset with general visual classes like ImageNet, this hints at employing different layers of the network to enable discovery at different levels of semantic.
Interestingly, the winning entry of ImageNet 2015, the so-called ResNet~\cite{kaiming16resnet}, substantially improves accuracy by introducing skip connections in CNN architectures.  However, for similar instance search, aggregation strategies~\cite{BL15} significantly outperform the choice~\cite{BSCL14} of simply extracting the activation at a given layer.

Works on co-segmentation~\cite{JBP12} and the approach of Cho~\etal.~\cite{CKSP15} aim at discovering objects by matching image regions. These techniques are accurate but do not scale beyond a few thousand images as they require maintaining and processing local descriptors. In contrast, we use only global image descriptors.

\paragraph{Clustering \& kNN Graph.}

The gold-standard clustering method is k-means. Min-hashing~\cite{ZJG13a} or binary k-means~\cite{GPYBBF15} have also been considered for visual discovery.
However algorithms that can take an arbitrary metric on input are more flexible.
We consider in particular clustering methods based on a diffusion process, which share some connections with spectral clustering~\cite{NLCG08}. %
They are an efficient way of clustering images given a matrix of input similarity, or a kNN graph, and have been successfully used in a semi-supervised discovery setup~\cite{FWT09}.
In~\cite{PZ08}, a kNN graph is clustered with spectral clustering, which amounts to computing the $k$ eigenvectors associated with the $k$ largest eigenvalues of the graph, and clustering these eigenvectors. Interestingly, when the eigenvalues are obtained via Lanczos iterations~\cite[Chapter~10]{GL13}, the basic operation is still a kind of diffusion process.

This is also related to Power Iteration Clustering~\cite{LC10}. In our experiments we evaluate a simplified version of it proposed by Cho \etal~\cite{CL12} to find clusters: instead of clustering a low-dim space, we follow the path to the mode of each cluster.
We refer the reader to \cite{DB13} for a review of diffusion processes and matrix normalizations.
Approximate algorithms~\cite{DCL11,KKNMS16,AKAE15,HD16} have been proposed to efficiently produce the kNN graph used as input of iterative/diffusion methods, some of them operating in the compressed domain. %

\paragraph{Similarity or distance normalization.}

In retrieval applications, images are typically ordered by distances, meaning that only the relative distances to the query matter. However, discovery is a detection problem, and its quality depends on the absolute distances between all pairs of descriptors. When building a kNN graph, it is therefore important to ensure that edges originating from different nodes have comparable weights. %
This problem is well known in spectral clustering~\cite{ZP04} and computer vision~\cite{PLR09,SKBB12}, and has led authors to propose different normalization pre-processing of distances or similarities. For instance, the contextual dissimilarity measure~\cite{JHS07} regularizes distances by local updates.
Another related work by Omercevic \etal~\cite{ODL07} uses the distribution of points relatively \emph{far away} from the current point to regularize the distance distribution.
This empirical choice is supported~\cite{FJ13} by extreme value theory and estimation, which was also been successful to calibrate the output of classifiers~\cite{SKBB12}. We use a simpler version of this regularization~\cite{JDS11a} and symmetrize it.

\section{Discovery pipeline}
\label{sec:method}

This section describes the different methods and choices involved in our discovery pipeline, namely the image description, kNN graph construction and metric normalization when applicable, and four clustering algorithms subsequently evaluated in Section~\ref{sec:experiments}.

\subsection{Description: combining semantic levels}
\label{sec:description}

The image descriptors must be (1) reasonably fast to compute, and (2) compact enough so that the clustering algorithms can handle them afterwards. For (1), we chose a 34-layer ResNet, trained on an unrelated image classification dataset as baseline descriptor.

\begin{figure}
\includegraphics[width=\linewidth]{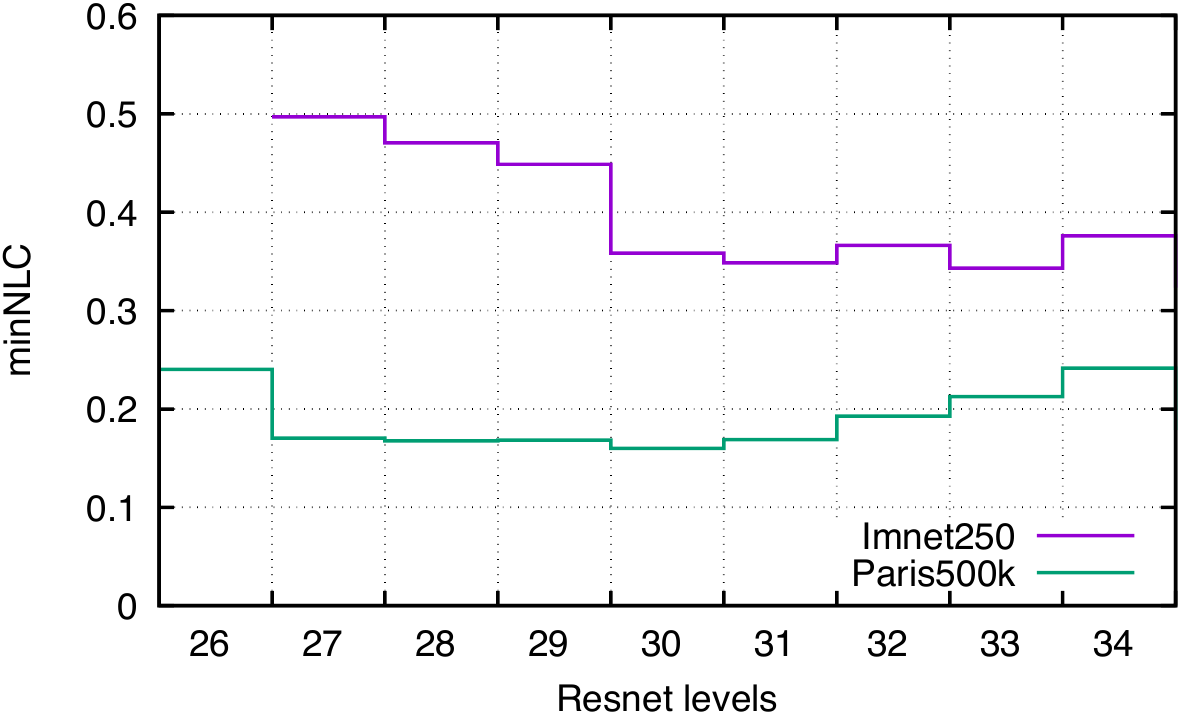}%
~\\
~\\
\scalebox{0.9}{
\begin{tabular}{|c|cc|c|c|}
\hline
              & \multicolumn{2}{c|}{single descriptor} & PCA128 & concatenation \\
              & level & performance & &  \\
\hline
ImageNet250   & high (33) & 0.376 & 0.353  & 0.323 \\
Paris500k     & low (30) &  0.160  & 0.180  & 0.179 \\
\hline
\end{tabular}}
\caption{\label{tab:pca}\label{fig:descs}
	\emph{Top}: discovery performance for k-means (minNLC, lower is better) as a function of the CNN activation level for the two evaluation datasets (ImageNet for classification and Paris500k for instance search). \emph{Bottom}: impact of PCA dimensionality reduction and concatenation.
	See Section~\ref{sec:experiments} for details on the datasets and the evaluation.
	}
\end{figure}

Figure~\ref{fig:descs} shows the clustering performance based on descriptors from several activation maps of the ResNet, for instance and classification tasks. When activation maps have a spatial extent (\ie, they are not 1x1 pixel), we aggregate them into a 512D descriptor using the \textbf{RMAC} technique~\cite{TSJ16}: this an aggregation of overlapping windows extracted from the map, whitened and L2-normalized. RMAC lays at the basis of many state-of-the-art methods for instance search~\cite{GARL16,RTC16} when applied to full-resolution images.

Given these results, we picked two 512D image descriptors:
\begin{itemize}
\item \textbf{high-level}: vector from the 33rd layer (just before the last fully connected layer).
\item \textbf{low-level}: the RMAC of the $7\times7\times512$ activation map of the 30th layer.
\end{itemize}

To make them more compact, the low- and high-level descriptors are both PCA-reduced to 128 dimensions, L2-normalized and concatenated. PCA dimensionality reduction is routinely adopted to process features extracted from neural networks~\cite{BSCL14,TSJ16}, and in fact PCA whitening is part of the RMAC aggregation.

The table in Figure~\ref{tab:pca} shows the impact of this choice. Starting from the full descriptor, the PCA from 512D to 128D has an impact of 2 points (negative for instance search, positive for classification). Concatenating the two descriptors \emph{improves the classification performance} significantly and has no impact on instance recognition. Therefore, in the following, we use a single concatenated description vector in 256D.

We also experimented by combining kNN graphs built separately from the low- and high-level features, but the resulting performance was at best identical to that of the concatenated features.

\subsection{kNN graph construction on the GPU}

Three of the four clustering algorithms we consider in this section use a matrix as input containing the similarity between all the images of the dataset. %
The \textbf{graph matrix} $A\in \mathbb{R}^{N\times N}$ is sparse and is equivalent to a kNN graph connecting each image to its neighbors, as determined by the similarity metric.

To construct the graph, we use a multi-GPU implementation of kNN search, implemented in the Faiss library\footnote{Available at \url{https://github.com/facebookresearch/faiss}.}~\cite{JDJ17}.
For small collections, \ie, up to 1 million images, we use a brute-force exact graph construction. For larger datasets, we use the Faiss IndexIVFFlat structure.
Some Faiss search methods operate in the compressed domain, but we do not use them because they are slower on the GPU. Besides, since the memory usage is dominated by the matrix storage, we do not benefit from compression.

\subsection{Clustering algorithms}

We now introduce four clustering methods that we evaluate for the discovery task. The first is a regular k-means applied on the input descriptor. The three other ones use as input the sparse similarity matrix $A$, post-normalized with metric normalization and symmetrized, which amounts to adding $A^\top$ to $A$. %
The best normalization strategy depends on the method, but it typically involves a bandwidth parameter that controls the importance of weak versus strong edges. %
The key features of the algorithms are summarized in Table~\ref{tab:algorithms}. %

\def \x {{x}}
\begin{table}
\begin{center}
{\small
\begin{tabular}{lcccc}

Algorithm
& \begin{minipage}[c]{1cm}use\\graph\end{minipage}
& \begin{minipage}[c]{1cm}update\\variable \end{minipage}
& \begin{minipage}[c]{1.2cm}hyper-parameter\end{minipage}
& \begin{minipage}[c]{1.2cm} runtime~(s) \end{minipage}
\smallskip \\
\hline
k-means   &     &  centroids    & k = 10000    & 21.3\\
AGC       &  \x &  node weights & $\tau=200000$& 21.4 + 0.24\\
PIC       &  \x &  node weights & $\sigma=0.5$ & 21.4 + 0.35\\
MCL       &  \x &  edge weights & $r=1.4$      & 21.4 + 44.6\\
\hline
\end{tabular}}
\end{center}
\caption{Summary of the evaluated algorithms and their typical runtimes on ImageNet250 (300k images). %
Each algorithm has a parameter that sets the granularity of the clusters, we indicate its optimal value. For the methods that build upon the kNN graph, the graph construction time is added. %
\label{tab:algorithms}}
\end{table}

\paragraph{K-means.}

We use the multi-GPU k-means implementation of Faiss. Performing a k-means on $N$\,=\,$100$ million descriptors is fast compared to the step of extracting the descriptors with a ResNet\footnote{The k-means complexity is determined by
$n_\textrm{iter} \times N \times k \times d$. With $n_\textrm{iter}=25$, $d=256$, $k=10^5$ and a dataset comprising $N=95$ million images, meaning about 640 Mflops per image. This figure should be compared to 3.6 Gflops reported for the ResNet architecture~\cite{kaiming16resnet}, and even more for the VGG network~\cite{SZ14}}.
Our multi-GPU implementation produces the clusters in about 15\,min with 8 Nvidia Titan X Maxwell GPUs, which we reduce to 4\,min by sub-sampling the descriptors during the E-M iterations.

\paragraph{Agglomerative Clustering (AGC)}

Agglomerative (or single-link) clustering depends only on the ordering of the edge weights. It removes edges that are below a given similarity threshold and identifies the connected components.
Therefore, the weights must be globally comparable and a normalization pre-processing step is important. A simple similarity normalization~\cite{JDS11a} that updates each similarity by subtracting from it a similarity to a far away neighbor (the rank-50 nearest-neighbor) works the best in practice. %

 When swiping over the thresholds, a binary tree is generated where each cluster is a node and the two children of a node are two clusters at a finer granularity that were fused to produce the node. Any number of clusters $\tau$ can be obtained by stopping the agglomeration at a given threshold.
A recent study~\cite{KML16} observes that such a single-link clustering tends to produce long chains. Our experiments in Section~\ref{sec:experiments} concur with this observation.

\paragraph{Power Iteration Clustering (PIC)}

Power iteration clustering finds a stationary distribution over the nodes of the graph by repeatedly multiplying a vector with the graph matrix until convergence. The actual clusters are typically extracted from the final distribution by clustering them in 1D~\cite{LC10}. However, this approach is hard to tune because it requires stopping the iterations before the clusters become indistinguishable. Therefore, we use a simple variant~\cite{CL12} where the clusters are identified by following the neighbors by a steepest ascent to a local maximum of the stationary distribution.
Similar to other works~\cite{CL12}, we found that a negative exponential to convert distances to weights $x \mapsto \exp(-x^2/\sigma^2)$ produces the best results, with $\sigma$ controlling the bandwidth.

\paragraph{Markov Clustering (MCL)}

This algorithm iterates over the similarity matrix as
\begin{align}
A \leftarrow A\times A \\
A \leftarrow \Gamma_r(A)
\end{align}
where $\Gamma_r$ is an element-wise raising to power $r$ of the matrix, followed by a column-wise normalization~\cite{EDO2002}. The power $r\in (1, 2]$ is the bandwidth parameter; when $r$ is high, small edges are reduced quickly along the iterations. A smaller $r$ preserves the edges longer. We found the matrix converges in 10-50 iterations. The clusters are read from the final matrix by extracting the connected components.

An important computational parameter is the sparsity of the matrix, determined by the number of non-zero elements of the matrix. After each $A\times A$ product, we use a global threshold on the matrix to force low elements to 0. If the matrix contains $kN$ non-zero elements, the storage and computational complexity of one iteration is $\mathcal{O}(Nk^2)$. Because of this storage requirement, MCL is only applicable to relatively small collections (million-sized). To normalize $A$, we linearly map the rows of $A$ to the $[0,1]$ interval.

\section{Experiments}
\label{sec:experiments}

This section describes our experiments carried out on the instance and category discovery tasks. %

\subsection{Datasets}

\begin{table}[t]
\scalebox{0.85}{
\begin{tabular}{lrrrr}
\hline
dataset     & \# images & \# labeled & \# classes & class size (min/max) \\
\hline
ImageNet250 & 319512    & 319512       & 250      & 860/1300\\
Paris500k   & 501356    & 94303        & 79       & 114/22799 \\
Flickr100M  & 95074575  & 0            & 0        & N/A \\
\hline
\end{tabular}}
\caption{
	The three image datasets. %
	\label{tab:datastats}
}
\end{table}

\begin{figure*}[t]
\includegraphics[width=0.48\linewidth]{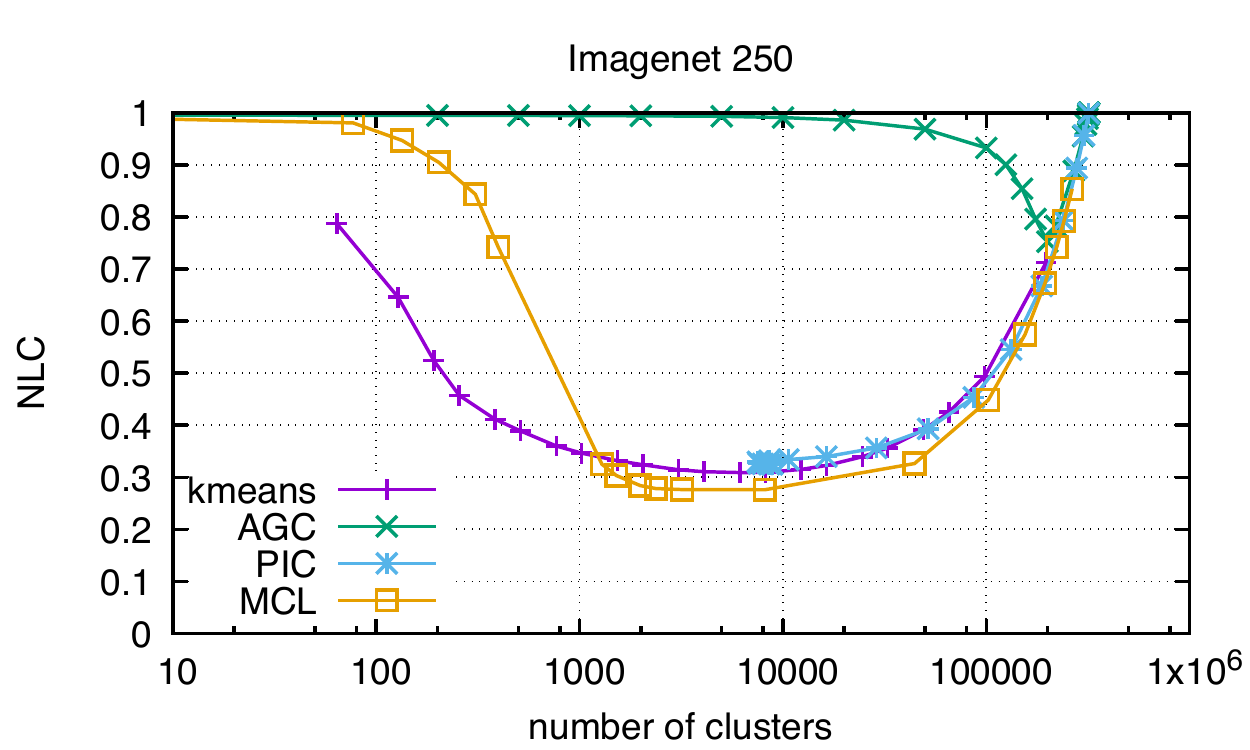}
\hfill
\includegraphics[width=0.48\linewidth]{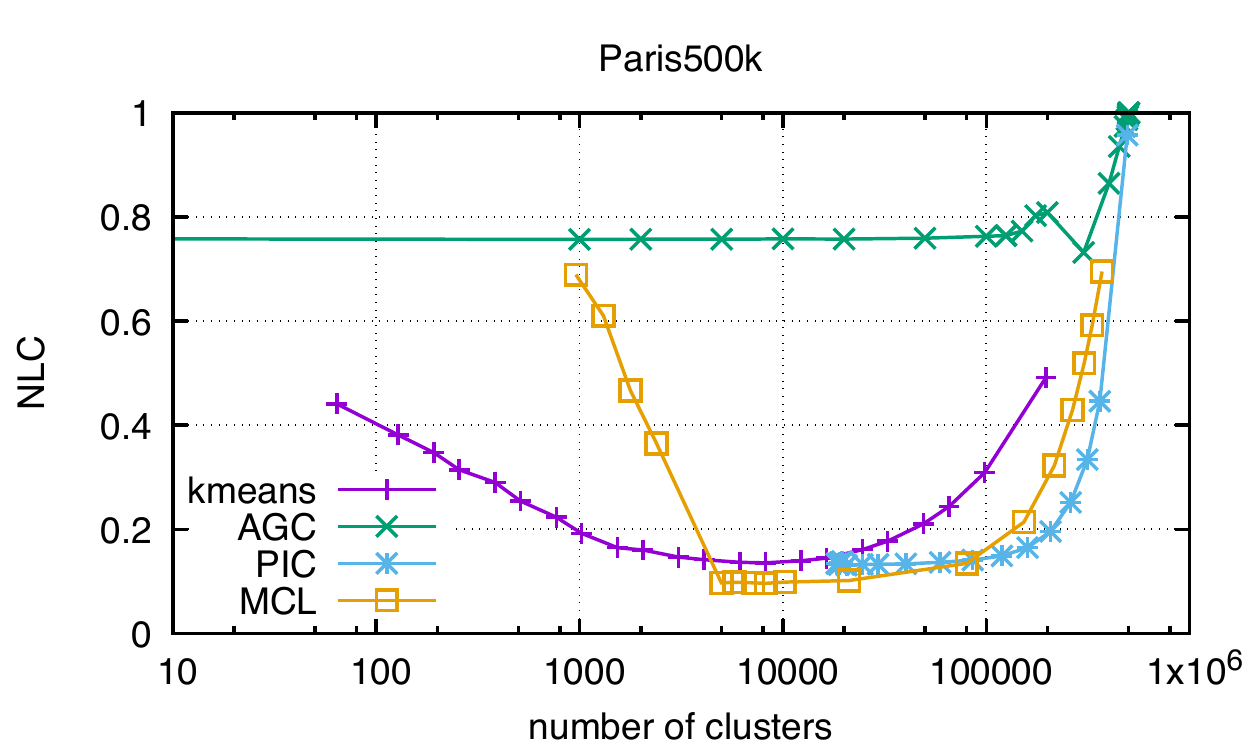}
\caption{\label{fig:clusterings}
	Comparison of clustering methods, in terms of NLC. By varying the hyper-parameters of Table~\ref{tab:algorithms}, the number of clusters (x axis) can be adjusted.
}
\end{figure*}

We use 3 datasets in this study, see Table~\ref{tab:datastats} for statistics.
\smallskip

\paragraph{ImageNet}
We use ImageNet~2012~\cite{DSLLF09} for evaluating the semantic discovery performance.
We withhold the images from 750 classes (chosen at random) out of the 1000 to train the ResNet image descriptor.
The ImageNet250 dataset is the set of classes that remain and used for evaluation of class discovery. The class sizes are balanced by design.
\medskip

\paragraph{Paris500k}
For the instance search dataset we use the Paris500k collection~\cite{WHL10}. It contains a set of Paris images from photo sharing sites, including landmarks, buildings, paintings, fa\c{c}ades of caf{\'e}s, etc.
The authors did an extensive study of this dataset~\cite{WL15}, with useful insights on the types of objects that appear in it, the
reliability of geometrical matching, how to find representative images, etc.
The dataset is partially labelled into classes, \ie{} the unlabelled part of the dataset \emph{also} contains instances of the classes.
\smallskip

\paragraph{YFCC100M}
This dataset~\cite{TFENPBL16} contains 100 million representative images from the Flickr photo sharing site (we managed to
download 95M of them). We use these images as \myemph{distractors} and consider them as unlabeled, even if some works have shown that the tags
or GPS metadata can be used as weak supervision~\cite{JMJV16,WKP16}. The images are diverse. A large fraction is portraits; there are also series of images from CCTV cameras.
\smallskip

\paragraph{Image description}
The two image descriptors we employ are described in Section~\ref{sec:description}.
We trained the ResNet on 750 classes\footnote{We used the resnet implementation from \url{https://github.com/facebook/fb.resnet.torch}} on
4 Nvidia K40 GPUs during 3 days. The final top-1 error after 90 epochs is 26.5~\%.
To analyze the images, we resize all images to $244\times244$ pixels and do a forward pass of the ResNet and keep activation maps of the layers we are interested in. Each minibatch of 128 images is processed in 670\,ms on a K40.
\smallskip

\paragraph{Dataset bias}
When combining datasets, it is important to be aware of the biases that define the datasets~\cite{TE11}. Some bias may cause the generation of dataset-uniform clusters, which makes the distractor set pointless. \textit{A priori}, all images are mined from similar photo sharing sites (Flickr and Panoramio), but a different sampling or image preprocessing may introduce some bias as well.

We observe such a bias on the Paris dataset: many generated clusters were suspiciously pure clusters from Paris500k.
To check this, we selected images from YFCC100M with the same selection criterion as Paris500k (on the GPS bounding box). Then we measured how the  retrieval mAP for the labelled part of Paris500k decreased when adding distractors from Paris500k and Paris images selected from YFCC100M. The mAP decreases similarly, which shows that the only bias is due to the semantic content of the images.

\subsection{Clustering performance evaluation}

Given a reference clustering, there are several clustering performance measures that
evaluate how similar the found clusters are to the ground truth classes (aka ``reference clustering'').
Classical measures include the normalized mutual information, cluster purity and rand index~\cite{LC10}.

\paragraph{Labelling cost, NLC and MinNLC}

In this work, we choose the \emph{labelling cost} (LC) as a performance measure. This cost was initially introduced
by Guillaumin~\etal{} for a face labelling task~\cite{GVS09}. It simulates the cost of an annotation interface that would be built
on the given clustering. The annotator sees the clusters one after another, and can take two possible actions:
(a) annotating the whole cluster of faces with a name, and (b) correcting the names of the faces of the cluster
that are not the dominant identity of the cluster.
The advantage of this measure is that it has a ``physical'' interpretation, and also offers an elegant
way of selecting the tradeoff between under- and over-segmentation of the dataset.
It is a cost, so lower is better.
It is bounded by the the number of classes (lower bound, reached with a perfect clustering)
and the number of images (upper bound, reached if each image gets a cluster).

To compare datasets of different sizes, we divide the LC by the number of images to annotate, yielding the
\myemph{normalized labelling cost} (NLC).
We often evaluate labelling costs for various clusterings that offer coarse-to-fine tradeoffs. In this
case we report the minimum NLC over all cluster sizes (minNLC).

\paragraph{Precision and recall.}

To compare with prior studies on the Paris dataset, we report the measures defined in the work by Weyand et al.~\cite{WHL10}, called \emph{precision} and \emph{recall} (somewhat misleadingly in a document retrieval context).
Here, \emph{precision} is computed as the number of images whose class is dominant in the cluster they are assigned to, normalized by the total number of images. This is related to cluster purity, but larger classes get a higher weight. The authors argue that this reflects
applications where larger classes are simply more important.
\emph{Recall} is the dual of precision; it is the fraction of images that belong to the cluster that contains most images of their class. Achieving a high recall means that the images of a given class are not spread out over several clusters.

\paragraph{Handling distractors.}

Distractors are unlabelled images that come from Paris500k and YFCC100M. They may or may not belong to one of
the classes we are evaluating the clustering on. For our NLC measure we follow the practice of
Weyand \etal\, and the ``junk'' images for Oxford Building evaluation~\cite{PCISZ07}: we \myemph{ignore the distractors in the computation of NLC}. The measures are still relevant, because
if many images with the same label are clustered together, it is likely that the unlabelled images of the cluster are also from the same class.

\begin{figure}
\includegraphics[width=\linewidth]{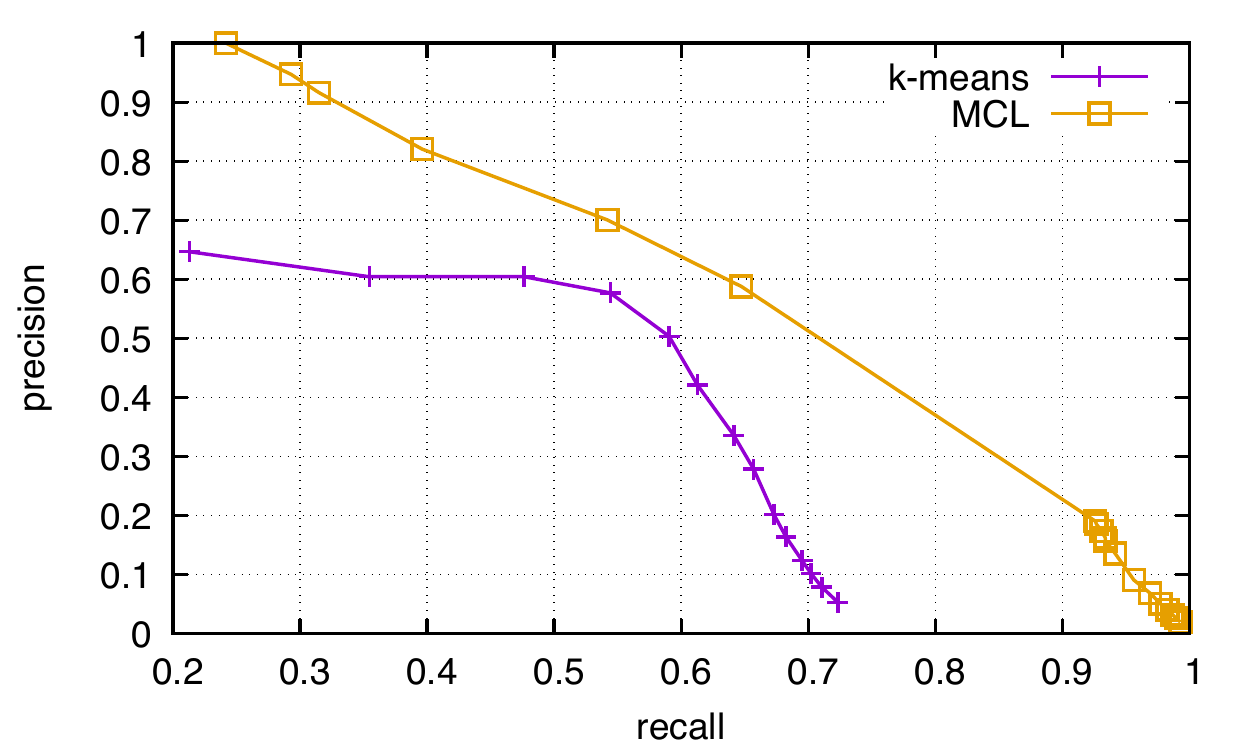}
\caption{\label{fig:prparis}
	Precision vs. recall on the Paris dataset.
}
\end{figure}

\subsection{Results on the individual datasets}

In Figure~\ref{fig:clusterings} we compare the clustering methods in terms of labelling cost, swiping different numbers of clusters.
The first observation is that the NLC for Paris500k
is much lower than that of ImageNet250, which reflects the fact that instance recognition
is an easier task than image classification, for typical modern datasets.
This is true despite the fact that the descriptors we use are close to the state of the art for image classification,
but quite sub-optimal for instance search, since the R-MAC descriptions are extracted at a
fixed resolution and without any fine-tuning of the convolutional part of the CNN~\cite{GARL16,RTC16}.

\paragraph{ImageNet250.}

The MCL method is the clear winner, followed by k-means and PIC, while AGC gives very poor performance.
The best performance is obtained for a number of clusters in between 1000 and 10000, which is larger than the number of categories of ImageNet250: it is easer for an annotator to label slightly over-segmented clusters than to dive into large clusters to individually label their contents.

\paragraph{Paris500k.}

The ranking of methods is about the same as for ImageNet250. Note that for this dataset, the largest class is that of the Eiffel Tower, the best strategy when presented with a single cluster of all images is to label them all as Eiffel Tower (which is correct for 22\% of the images), and correct the remaining images. This explains that the NLC is bounded at 0.78 for low numbers of clusters.

The clustering P-R is the standard performance measure for this dataset, and allows a direct comparison %
to previous studies. The performance that we achieve is lower than that reported in the original paper~\cite{WHL10}, which is expected since they use a full geometrical method and require a much more dense and costly comparison method. %

The comparison to the results of Avrithis \etal~\cite{AKAE15}, which considers a more similar setup and is oriented towards efficient discovery, shows that our method obtains much better results (they have P-R operating points of around $(0.42, 0.10)$). This is partly because we use a more powerful representation (ResNet rather than AlexNet), but also because our clustering method is better. More specifically, Figure~\ref{fig:prparis} shows that MCL is significantly better than k-means in this instance discovery scenario. Our method is also faster,  thanks to our better CPU and GPU implementations.

\subsection{Balanced clusters}

We analyze whether the four clustering methods produce balanced clusters in terms of size.
Our measurements are carried on the ImageNet250 dataset, for which all classes have
a very similar number of images. We would therefore have expected the different methods to produce balanced clusters.

\begin{figure}
\includegraphics[width=\linewidth]{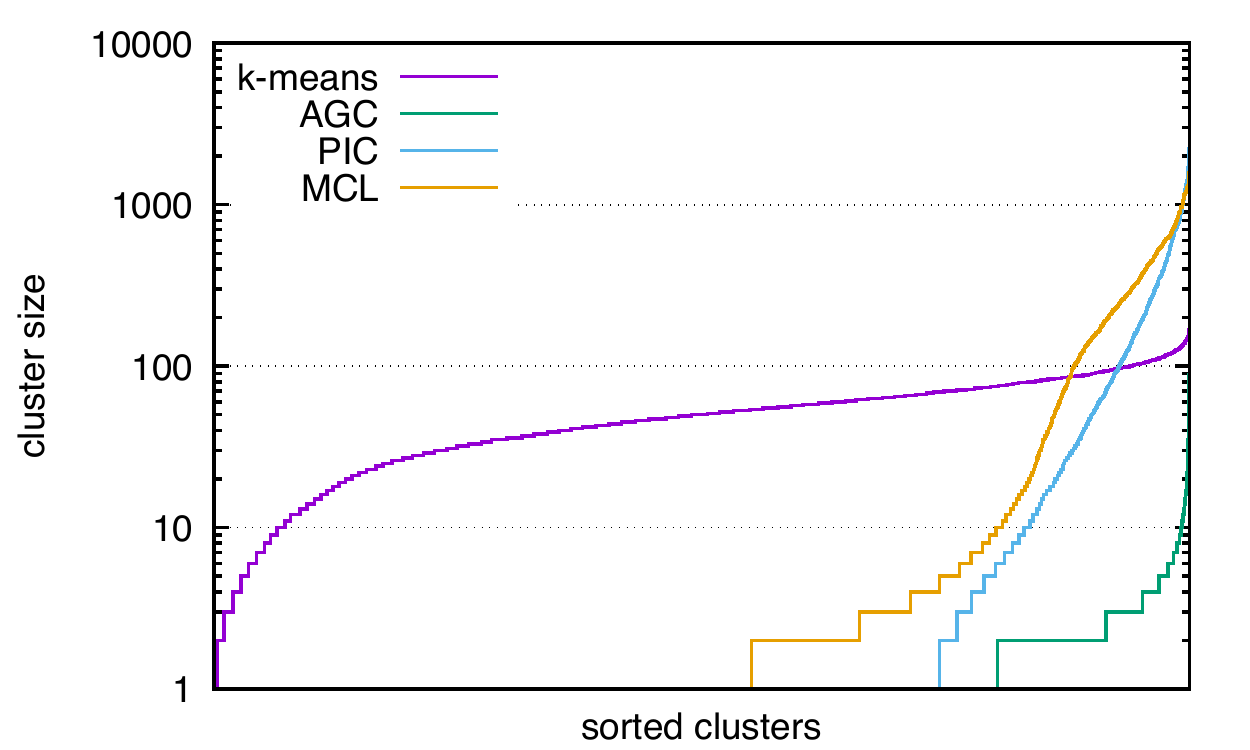}%
\caption{\label{fig:clustersizes}
	Sizes of the clusters produced by the clustering methods on ImageNet250, sorted from smallest to largest. The size of each clustering is chosen at the point where the minimum Labelling Cost is obtained.
}
\end{figure}

It is in fact not the case: Figure~\ref{fig:clustersizes} shows that k-means produces the most balanced clusters. For PIC and MCL about half of the clusters are singletons. The most unbalanced clustering is the agglomerative method. Its optimal operating point is at 200,000 clusters, which entails that 80~\% of its clusters are singletons.

\subsection{Large-scale results}

\begin{figure}
\includegraphics[width=\linewidth]{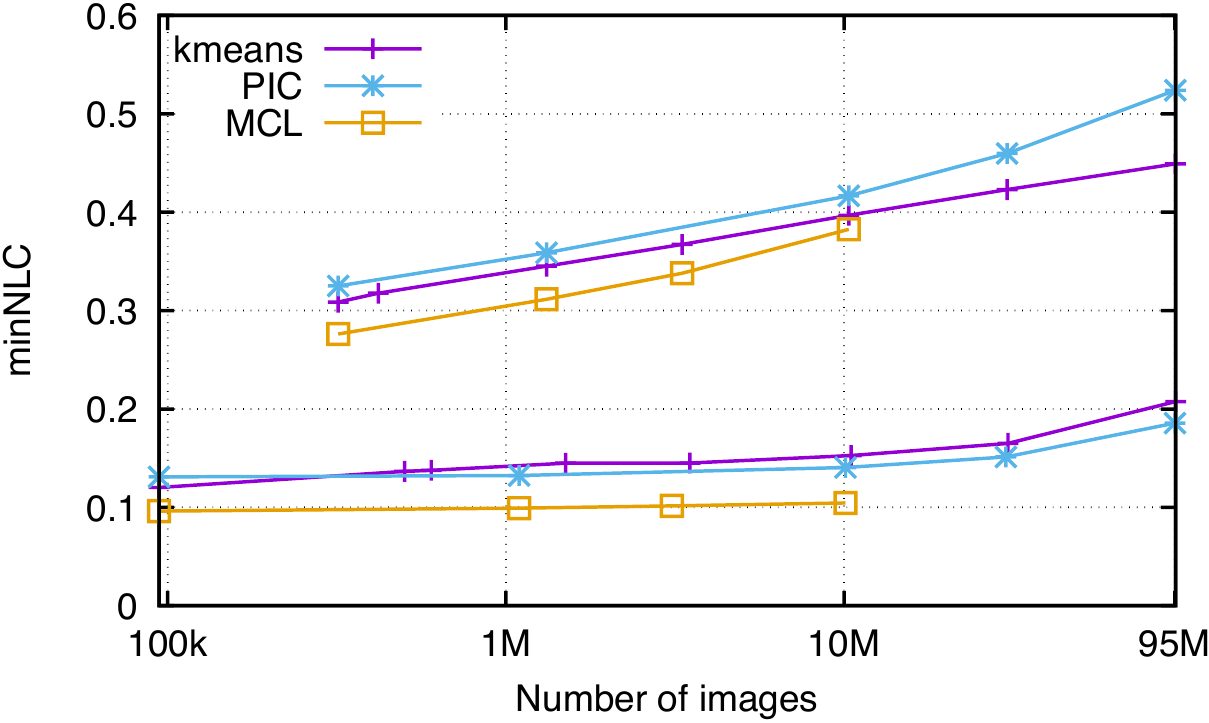}%
\caption{\label{fig:distractors}
	Clustering performance (minNLC, lower is better) as a function of number of distractors. The three curves above are for ImageNet250, the three below for Paris500k.
}
\end{figure}

We combine ImageNet250 and Paris500k with a varying number of distractor images to evaluate the performance of the discovery on a large scale.
Figure~\ref{fig:distractors} reports the performance as a function of the dataset size. We do not experiment with AGC, which is clearly inferior. MCL is difficult to scale beyond 10M images: the squared matrix $A\times A$ has up to 13~billion edges, and the total memory usage is up to 120~GB.
As expected, the performance degrades when the number of distractors increases. However, it degrades significantly slower for instance-level discovery than for class discovery. This is because the clusters have much clearer boundaries in the instance search case. In particular, MCL is almost not affected by distractors.

\subsection{Visual results}

\begin{figure*}
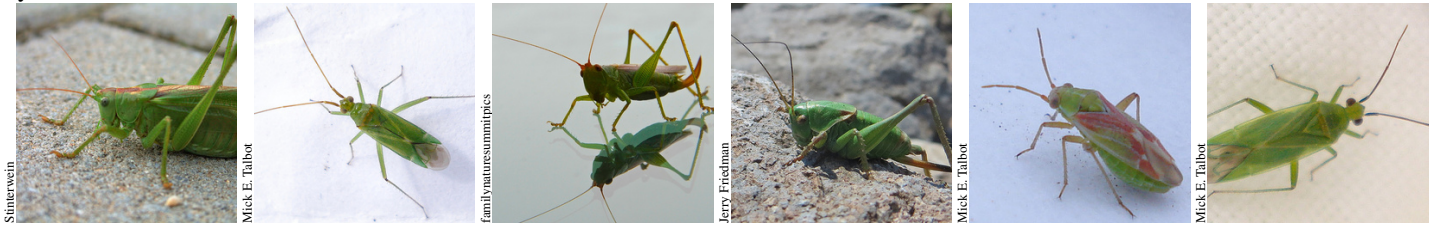

\input{figs/clusters/10151_e.tex.user}\\
\input{figs/clusters/69_e.tex.user}\\
\input{figs/clusters/6567_e.tex.user}\\
\input{figs/clusters/19921_e.tex.user}\\
\input{figs/clusters/28046_e.tex.user}\\
\input{figs/clusters/28404_e.tex.user}\\
\caption{\label{fig:clA}
	Example clusters with large intersections with a ground-truth category. For each cluster we indicate its size, the most frequent words from the Flickr annotations and the name of the ImageNet250 cluster with which it has the largest intersection. Although clusters contain Imagenet and Flickr500k images, we show only Flickr images for copyright reasons (and indicate the author's name).
}
\end{figure*}

\begin{figure*}
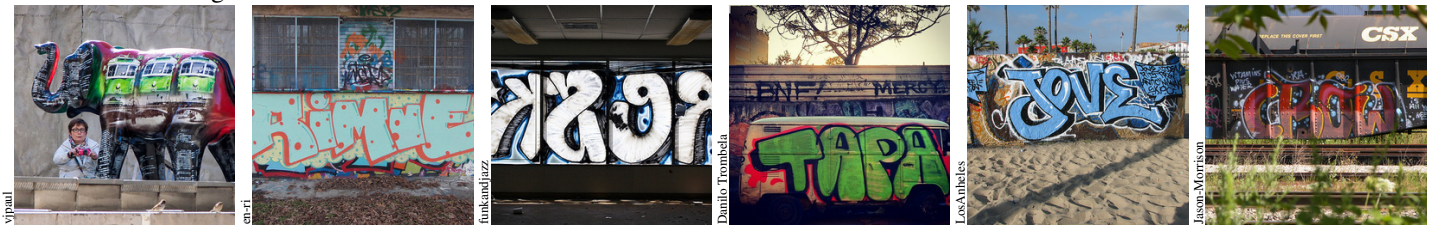

\input{figs/clusters/15313_e.tex.user}\\
\input{figs/clusters/20819_e.tex.user}\\
\input{figs/clusters/22088_e.tex.user}\\
\input{figs/clusters/3896_e.tex.user}\\
\input{figs/clusters/7362_e.tex.user}\\
\input{figs/clusters/9957_e.tex.user}\\
\caption{\label{fig:clB}
	Example clusters without any specific intersection with a ground-truth category.
}
\end{figure*}

We present examples of image clusters in Figures~\ref{fig:clA} and~\ref{fig:clB}. The clusters are obtained by mixing both ImageNet250 and Paris500k with 95M images from YFCC100M. Recall that we rely on the visual content only to produce the clusters. %
To get an idea of how this could be combined with image tags to automatically label the clusters, we report the available annotations for the clusters: for ImageNet250 this is the synset name. For YFCC100M, we construct a bag of words (BoW) from the captions of the images of each cluster and report the most frequent words. The classes of Paris500k are not labelled.

Figure~\ref{fig:clA} shows that it is possible to \myemph{propagate} the ImageNet250 annotations to a whole cluster, or to find a more accurate name for animal species (dog $\leftrightarrow$ Bedlington terrier). For the Paris500k images, the BoW annotation gives a reliable name for the locations viewed in the images.

Figure~\ref{fig:clB} shows that there are many new clusters that also appear in the dataset. They are typically related to events (prom, concert), to objects that are not in the ImageNet collection (grafitti, fashion), or to combination of several classes occurring simultaneously in the cluster's images.

\newcommand{\ig}[1]{\includegraphics[width=2.9cm]{figs/clusters/#1}}
\newcommand{\igcombine}[3]{\rotatebox{90}{\tiny{\rule{0pt}{1em}#3}}\makebox[0pt]{\hspace*{2.9cm}\ig{#1}}\ig{#2}\,}

\newcommand{\IGimnetsolid}[1]{\igcombine{#1}{red_solid.pdf}{}}
\newcommand{\IGparisdashed}[1]{\igcombine{#1}{blue_dashes.pdf}{}}
\newcommand{\IGparissolid}[1]{\igcombine{#1}{blue_solid.pdf}{}}

\newcommand{\IGflickrdashed}[2]{\rotatebox{90}{\tiny{\rule[-0.2em]{0pt}{1em}#2}}\ig{#1}\,}

\section{Conclusion}
\label{sec:conclusion}

This paper presents a thorough evaluation of a large-scale discovery pipeline for both visual instances and categories.
Our analysis of different clustering methods, distance normalizations, and descriptors shows that the best choices depend on the scale of the problem. The Markov Clustering algorithm offers the best quality but is scale-bounded because of the size of the affinity matrix.
For large collections such as the YFCC100M dataset, Power Iteration Clustering and k-means are the best competitors.

Our experiments have been carried out with the novel and efficient multi-GPU implementations of the Faiss library, typically able to cluster 95 million images into 100,000 groups on one machine in less than 5 minutes. As a result, we report state-of-the-art results with respect to the trade-off between performance and efficiency.

Another conclusion is that category-level clusters can be improved by using lower-level descriptors. We plan to publish code and data that reproduce the experiments.

\newpage

{%
\bibliographystyle{plain}
\bibliography{egbib}
}

\end{document}